\documentclass[conference,10pt]{IEEEtran} 
\usepackage{cite}
\usepackage{amsmath,amssymb,amsfonts}
\usepackage{graphicx}
\usepackage{multirow}
\usepackage{textcomp}
\usepackage{xcolor}
\usepackage{booktabs}
\usepackage{array}
\usepackage{subfigure}
\usepackage[linesnumbered,ruled,vlined,algo2e]{algorithm2e}
\usepackage[marginal]{footmisc}
\DeclareGraphicsExtensions{.eps}
\begin{document}
\title{\huge{
Emergency Computing: An Adaptive Collaborative Inference Method Based on Hierarchical Reinforcement Learning
}\vspace{-0.5cm}}

\author{\IEEEauthorblockN{Weiqi Fu\IEEEauthorrefmark{1},
Lianming Xu\IEEEauthorrefmark{2},  Xin Wu\IEEEauthorrefmark{1}, Li Wang\IEEEauthorrefmark{1}, and Aiguo Fei\IEEEauthorrefmark{1}
\vspace{-0.2cm}
}\\
\small\centerline{\IEEEauthorrefmark{1}School of Computer Science (National Pilot Software Engineering School), }\\ {Beijing University of Posts and Telecommunications, Beijing, China}\\\IEEEauthorrefmark{2}
School of Electronic Engineering, Beijing University of Posts and Telecommunications, Beijing, China\\
Email:{\{fuweiqi, xulianming, xin.wu, liwang, aiguofei\}@bupt.edu.cn}\\
\vspace{-1cm}

        }\vspace{-0.5cm}
\maketitle

\begin{abstract}
In achieving effective emergency response, the timely acquisition of environmental information, seamless command data transmission, and prompt decision-making are crucial. This necessitates the establishment of a resilient emergency communication dedicated network, capable of providing communication and sensing services even in the absence of basic infrastructure. In this paper, we propose an Emergency Network with Sensing, Communication, Computation, Caching, and Intelligence (E-SC3I). The framework incorporates mechanisms for emergency computing, caching, integrated communication and sensing, and intelligence empowerment. E-SC3I ensures rapid access to a large user base, reliable data transmission over unstable links, and dynamic network deployment in a changing environment. {\color{black}However, these advantages come at the cost of significant computation overhead. Therefore, we specifically concentrate on emergency computing and propose an adaptive collaborative inference method (ACIM) based on hierarchical reinforcement learning.} Experimental results demonstrate our method's ability to achieve rapid inference of AI models with constrained computational and communication resources.

\end{abstract}
\thispagestyle{empty}

\renewcommand{\thefootnote}{}
\footnote{\indent {\color{black}This work was supported in part by the National Natural Science Foundation of China under grants U2066201, 62171054, 62101045, and 62201071, in part by the Natural Science Foundation of Beijing Municipality under Grant L222041, in part by the Fundamental Research Funds for the Central Universities under Grant No. 24820232023YQTD01, and 2023RC96, in part by the Double First-Class Interdisciplinary Team Project Funds 2023SYLTD06. (Corresponding author: Li Wang.)}}

\section{Introduction}

Ensuring timely and effective emergency response is paramount for safeguarding lives and property during natural disasters like earthquakes, fires, and floods. Unfortunately, these events often lead to infrastructure loss, rendering public networks inaccessible and vehicle-mounted communication devices impractical. In such scenarios, rescue personnel must rely on mobile portable devices to establish temporary dedicated networks for command and communication services. However, the rescue environment is inherently complex and dynamic, fraught with unpredictable dangers and secondary disasters. The ability of rescue personnel to make swift and accurate decisions directly influences the success of the operation and the safety of the rescuers. The critical factor lies in rapidly obtaining environmental information and ensuring the transmission of command information. This necessitates an emergency communication dedicated network capable of supporting communication services such as massive Machine Type Communications (mMTC), Ultra-Reliable Low Latency Communications (URLLC), enhanced Mobile Broadband (eMBB), and providing sensing capabilities such as high-precision target detection or recognition.


\begin{figure*}[t]
    \centering
    \includegraphics[width=0.85\textwidth]{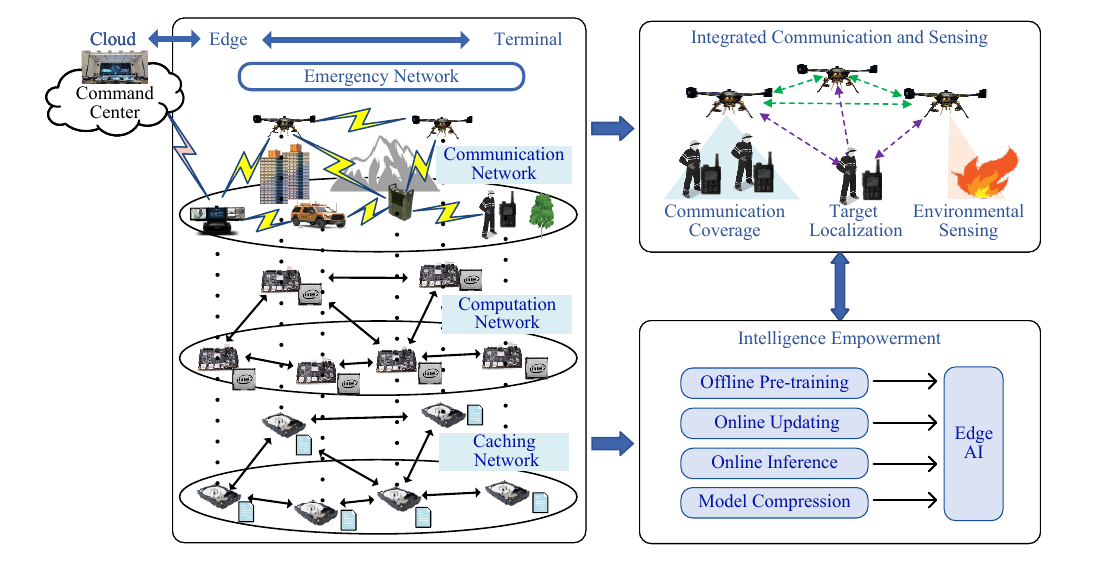}
    \caption{An Emergency Network with Sensing, Communication, Computation, Caching, and Intelligence.}
    \label{fig01}
    \vspace{-0.5cm}
\end{figure*}

However, the current design of emergency communication dedicated networks, which primarily considers a single communication resource for ensuring communication services, faces the following issues: Firstly, the limited spectrum resources allocated to emergency dedicated networks pose a significant issue, especially when serving diverse communication needs for potentially thousands of rescue team members. While technologies such as D2D and NOMA\cite{b2} can enhance spectrum efficiency, the swift completion of spectrum allocation remains challenging, leading to slow access. Secondly, existing emergency dedicated networks are susceptible to interference from terrain features such as trees, buildings, and mountains. The high mobility of personnel further contributes to unstable transmission links, potentially interrupting the transfer of large files such as videos and maps.
Lastly, the topology of existing emergency dedicated networks is relatively fixed, lacking adaptability to the dynamic nature of emergency scenes. This inflexibility results in link failures, leading to communication loss with rescue personnel. Consequently, the challenge at hand is how to establish a rapid, reliable, and sustained emergency communication dedicated network with limited resources. This calls for a deeper exploration of the collaborative capabilities of communication, computation, and caching resources\cite{b3} among network nodes, ensuring simultaneous communication and sensing services based on communication resources. {\color{black}Especially in terms of computational resources, all methods of resource allocation, network deployment, as well as communication and sensing services, rely on it, yet computational resources are constrained in emergency scenarios.}


In summary, the deployment of emergency dedicated networks in traditional scenarios often relies on manual experience or traditional optimization methods, making the decision-making process complex. Specific tasks require precise modeling, resulting in poor generality. With the development of AI technology, data-driven deep neural networks can effectively reduce the complexity of online decision-making models by combining offline pre-training, online updating, and online inference. These deep neural networks offer high scalability, good stability, and strong generality, enabling rapid intelligent decision-making in dynamic emergency command communication networks.


In response to these challenges, this paper proposes the E-SC3I framework, which extends traditional communication networks by incorporating enhanced computational and caching capabilities. Through the establishment of emergency computing and emergency caching mechanisms, 
this augmentation not only improves communication services but also transforms the network into a multi-objective one,  seamlessly integrating communication and sensing. {\color{black}Especially in the context of emergency computing}, the key to accomplishing time-sensitive emergency tasks like disaster detection, data transmission and topology optimization under limited resources and unreliable networks is to efficiently schedule multi-node communication and computational resources for low-latency collaborative computing. And leveraging AI technology, E-SC3I {\color{black}can} achieve real-time dynamic network deployment through communication-assisted sensing and sensing-guided communication. Thus, the E-SC3I framework adapts to time-varying emergency environments, facilitating rapid user access, reliable data transmission, and dynamic network topology optimization, while delivering intelligent communication and sensing services. The contributions of this paper are as follows:

\begin{itemize}
\item The E-SC3I framework is proposed for the first time, establishing four mechanisms: emergency computing, emergency caching, integrated sensing, and intelligence empowerment. This framework creates a rapid, reliable, and sustainable dedicated emergency communication and sensing network that can adapt to time-varying emergency environments.
\item In the context of emergency computing,  {\color{black}we propose the ACIM method,} introducing hierarchical reinforcement learning to address the pruning and partitioning problems in the collaborative inference of AI models. We design the state, action, and reward to achieve coupled optimization of pruning and partitioning strategies.
\item Experimental results show that, with the same accuracy, our proposed ACIM method can reduce the latency by up to $32\%$ compared to the 2-step pruning method \cite{b13}.
\end{itemize}

\section{E-SC3I Framework}
As shown in Fig.~\ref{fig01}, the E-SC3I framework is grounded in conventional air-ground emergency communication networks, establishing an interconnected network structure encompassing communication, computing, and caching. Through the coordination of communication-computation and communication-caching resources, E-SC3I institutes mechanisms for emergency computing and caching, facilitating the scheduling of multidimensional C3 resources. Upon this foundation, sensors can expeditiously transmit the gathered data through the communication network, facilitating its prompt availability for tasks such as environmental monitoring and target localization. The real-time data feedback from the sensing system serves as a guiding factor for the optimization and adjustment of the communication network. This process culminates in the realization of integrated sensing and communication. Ultimately, through the infusion of AI technology, E-SC3I amplifies communication and sensing capabilities, culminating in expeditious decision-making for the dynamic deployment of emergency networks.


\subsection{Emergency Computing}
To facilitate the dynamic deployment of the emergency network and the integrated communication-sensing multi-objective services like data compression and image recognition, sufficient computational resources are essential. The constrained computational capabilities of portable devices, coupled with the unpredictable nature of communication links to the cloud server in the command center, render conventional cloud computing architectures reliant on public network communication unsuitable in this situation. Therefore, there emerges a requirement for a cloud-edge-terminal collaborative computing architecture \cite{b17}. In situations of unstable communication links and insufficient computation resources, this architecture jointly optimizes transmission and computation latency. It establishes serial collaborative computing mechanisms among nodes with different computational resources and parallel collaborative computing mechanisms among nodes with equivalent computational resources. This design allows for low-latency computation on multiple nodes with limited resources, supporting the efficient functionality of the E-SC3I framework.

\subsection{Emergency Caching}
Unstable emergency network links lead to interruptions during the transmission of large files such as videos and maps. Emergency caching introduces coded caching technology\cite{b5}, which can adaptively complete the segmented coding of large files based on the transmission success probability of multiple links and achieve collaborative forwarding. This not only enhances the reliability of large file transmission but also reduces the bandwidth requirements of individual links. Furthermore, dynamic allocation mechanisms for computation and caching resources can be established based on the priority of communication and sensing services, spatial distribution, data frequency, and other factors, ensuring the reliability of the E-SC3I network.

\subsection{Integrated Communication and Sensing}

The optimization of emergency response efficiency hinges on the interplay between communication and sensing. In a time-varying emergency environment, the real-time adjustment of network node deployment relies heavily on environmental sensing. Simultaneously, environmental sensing is contingent upon a rapid, reliable, and sustained communication network. Consequently, through the integration of considerations such as communication rate, coverage area, transmission delay, and sensing accuracy, an integrated communication and sensing network can be established. Based on the integrated network, we can further gather multi-modal data such as infrared data and radar data, which are applied to the rapid construction and dynamic updating of on-site three-dimensional maps, providing a basis for rapid and accurate network dynamic deployment in complex and time-varying emergency environments, thereby ensuring continuous communication and sensing services.


\subsection{Intelligence Empowerment}
Establishing an E-SC3I network that coordinates multiple resources for communication, computation, and caching as well as multiple objectives for communication and sensing, inherently increases the complexity of network deployment decision-making. Leveraging AI technology to facilitate emergency computing, and caching, and extending its application to the communication and sensing services of the E-SC3I network holds the potential to enhance the network's adaptive capabilities. By effectively combining various technologies\cite{b6} such as offline pre-training, online updating and inference, and model compression, it is possible to achieve rapid deployment optimization of the emergency network for time-varying environments, resources, and services.

\begin{figure}[t]
    \centering
    \includegraphics[width = 0.49\textwidth]{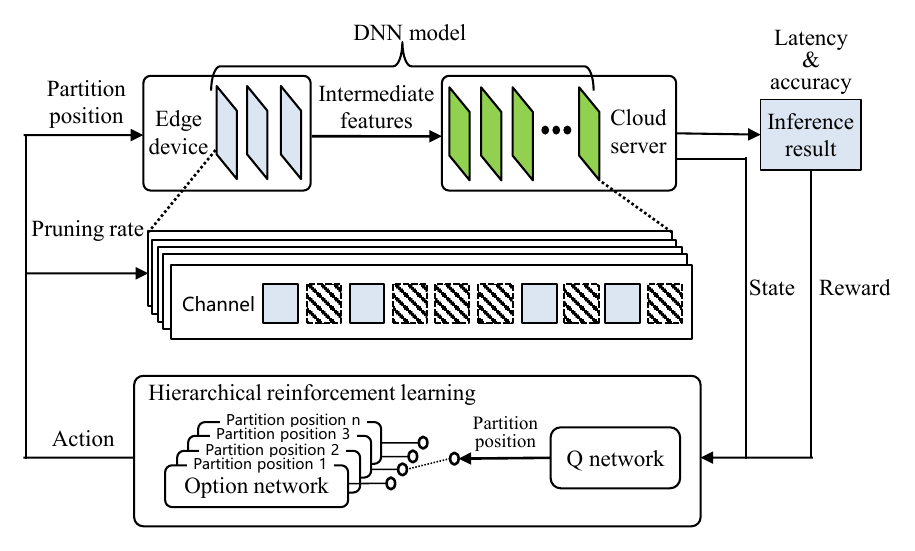}
    \caption{An Adaptive Collaborative Inference Method for Emergency Computing.}
    \label{fig02}
\vspace{-0.3cm}
\end{figure}

\section{An Adaptive Collaborative Inference Method}
The E-SC3I framework is capable of supporting computation-intensive services, such as object recognition, in resource-constrained emergency environments. The technical challenge lies in how to utilize limited computational and communication resources to provide low-latency, high-reliability intelligent services. Strategies for deep neural network (DNN) model compression, such as pruning \cite{b8}, distillation \cite{b11}, and quantization \cite{b12} have proven effective in reducing model parameter size and decreasing the computational cost of model inference. In the context of collaborative inference \cite{b14,b15,b16}, the model will be compressed and partitioned, undergo initial inference at the edge, and have intermediate results sent to the cloud to complete the remaining inference process. However, existing methods typically prune the model before partitioning, and the pruning process does not take into account the impact of model partition positions on computation latency and transmission latency. Yet, pruning strategies and partitioning strategies are interdependent. To address this, this paper proposes an adaptive collaborative inference method for emergency computing, {\color{black}called ACIM}, jointly determining the partition position and the pruning rates of an object recognition model under given accuracy constraints.

We assume that the pruning rate for each convolutional layer needs to be determined sequentially, following the order of the convolutional layers. Each decision only considers the current state of the object recognition model and environmental information. The decision-making process has Markovian property. Thus, the decision-making process of the pruning rate can be defined as a Markov decision process. In this process, the state includes the current state of the object recognition model itself, as well as environmental information such as transmission rate, computational power, and accuracy constraints. The action is a pruning rate of one convolutional layer, and the reward is determined by the inference latency and accuracy. Furthermore, considering that the decision of the partition position is made for the entire recognition model, while the decision of the pruning rate is made for a single convolutional layer, we define the decision of the pruning rate as sub-problems under different partition positions. These sub-problems can be simultaneously solved using the hierarchical reinforcement learning method.

\subsection{State}
We define the state in the decision-making process using 8 features, as follows:
\begin{equation}
\begin{split}s_t=\ \left\{R_{tran},{R}_{comp},{Acc}_{req},{FLOPs}_t,\right.\\
\left.{Channel}_t,{Data}_t,{Layer}_t,{Action}_t\right\}\end{split},
\end{equation}
where $R_{tran}$ is the transmission rate and $R_{comp}$ {\color{black}is the ratio of latency when inferring a model segment on the edge compared to on the cloud.}. ${Acc}_{req}$ stands for the minimum required inference accuracy based on the actual requirement. These parameters, $R_{tran}$, $R_{comp}$, and ${Acc}_{req}$, remain fixed as per the given scenario. To expedite the training process, we opt to utilize the accuracy of the model without resorting to fine-tuning. Here, ${FLOPs}_t$\ denotes the FLOPs of every layer during each decision step. ${Channel}_t$ is the number of channels in each convolutional layer during each decision step. The state of the convolutional layer before the partition position is encapsulated in ${Data}_t$, which includes FLOPs, the number of channels, and the size of output features. ${Layer}_t$ is represented as a one-hot array, where a value of 1 indicates the position of the layer slated for pruning.



\subsection{Action}
The decision-making process requires two actions. $a_{option}$ represents the partition position of the model, while $a_{ratio}$ represents the pruning rate of one convolutional layer. The hierarchical reinforcement learning method first outputs the action $a_{option}$ and then sequentially outputs the pruning rate $a_{ratio}$ for each convolutional layer in each decision-making step.

\subsection{Reward}
Due to the significant time cost involved in evaluating the inference accuracy and latency of the pruned recognition model, these measurements are not performed. Once the pruning ratio of the last convolutional layer is outputted, the inference accuracy and latency of the model are measured, and the reward is then calculated. The reward is defined as:
\begin{equation}
Reward=\left\{\begin{matrix}\frac{1}{\left(T_{edge}+T_{trans}+T_{cloud}\right)},&Acc\geq{Acc}_{req}\\0,&Acc<{Acc}_{req}\end{matrix}\right.,
\end{equation}
where $T_{trans}$ represent the transmission latency. $T_{edge}$ and $T_{cloud}$ represent the inference latency, which is measured by actual execution on a cloud server. The measured time is multiplied by the factor of $R_{comp}$ to indicate the inference latency on the edge.

\subsection{Loss Function}
As shown in Fig.~\ref{fig02}, the method consists of a Q-network and multiple option networks. The Q-network is responsible for predicting the cumulative reward obtained from the decision process, which is calculated based on the inference latency and accuracy. With the cumulative reward predictions of each partition position, the position with the highest predicted value will be selected. The loss function of the Q-network is defined as follows:
\begin{equation}
{Loss}_Q\ =\ \frac{1}{T}\sum_{t}\frac{1}{2}\left(Q_t-Return\right)^2
\end{equation}
{\color{black}where $Return$ is the cumulative rewards and $Q_t$ is the predicted cumulative rewards outputted by the Q-network.}

The option networks are responsible for determining the pruning rate of each convolutional layer. Each option network corresponds to a partition position. Once the Q-network completes the partition position decision, the pruning rate decisions for all convolutional layers are made by the corresponding option network. The loss function of option networks is defined as follows:
\begin{equation}
{\rm Loss}_{option}\ =\ \frac{1}{T}\sum_{t}\ Q_t
\end{equation}

\begin{figure}[!t]
    \centering
    \includegraphics[width=0.48\textwidth]{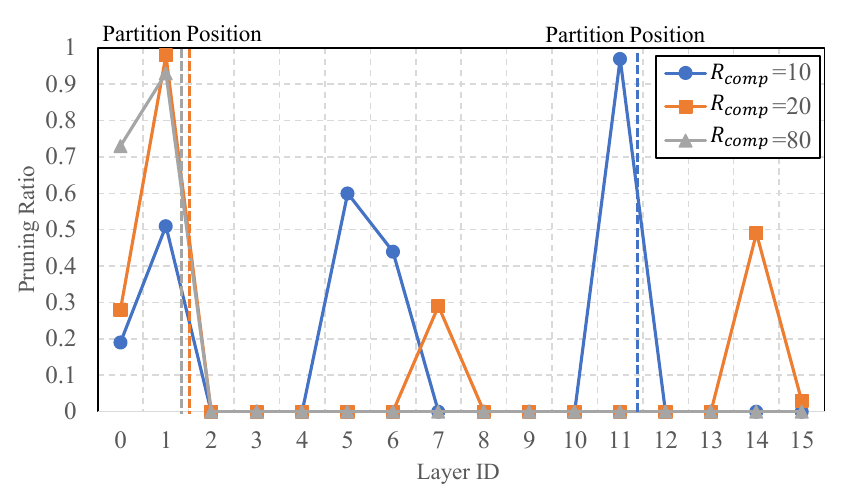}
    \caption{Pruning Strategies with Different $R_{comp}$
    .}
    \label{fig03}
\vspace{-0.3cm}
\end{figure}

\section{Experiment}
\subsection{Experimental Setup}

The experimentation utilized {\color{black}pretrained} recognition models, namely VGG16, VGG19, and ResNet34, implemented on the PaddlePaddle deep learning framework. Parameters $R_{tran}$ and $R_{comp}$ were set to 1280KB/s and 20, respectively. The experiment assesses the proposed method using the CIFAR-10 image classification dataset, consisting of images across 10 different classes commonly utilized for image recognition and classification tasks. The testing server was equipped with an 11GB RTX 2080 Ti GPU card and a 2.5GHz Intel Xeon Platinum 8255C CPU.

Inference accuracy and latency of pruned and partitioned models were assessed on the CIFAR-10 dataset with images sized at $320\times 320\times 3$. Both the Q-network and option networks in the hierarchical reinforcement learning method employed two hidden layers with 300 nodes each and a batch size of 128. The Q-network's learning rate was set to 1e-3, and the option network's to 1e-4. The soft updating parameter was 0.01. The priority experience replay memory had a size of 400n. Initially, for each partition position, 100 samples were pruned at a randomly generated rate. Subsequently, the partition position with the highest predicted reward was chosen, and the pruning rate was determined using an initial noise coefficient of 0.9 and a decay coefficient of 0.995.


\subsection{Performance Analysis}

In Fig.~\ref{fig03}, we compare the impact of different $R_{comp}$ values on decision outcomes under the same transmission rate and accuracy constraint. A lower $R_{comp}$ signifies more abundant edge computational resources. Experimental results demonstrate that with $R_{comp}$ set to 10, the partition position tends to be selected towards later positions. This is because, in later model layers, network layers produce smaller output features, resulting in reduced transmission latency. However, when $R_{comp}$ is set to 20 and 80, the partition position tends to be selected towards the earlier positions, effectively reducing computational latency. Regarding the pruning strategy, experimental results indicate that network layers before the partition position are more likely to be assigned higher pruning rates, especially the layer just before the partition position, which is assigned the highest pruning rate to substantially reduce transmission latency. Therefore, it can be concluded that the decision results have good adaptability to differentiated environmental factors.

\begin{table}[!t]
\centering
\caption{{\color{black}Comparison With 2-Step Pruning and L1-Normed Based Pruning}}
\resizebox{0.9\columnwidth}{!}{
\begin{tabular}{cccccc}
\toprule 
\multirow{2}{*}{\textbf{Model}} & \multirow{2}{*}{\textbf{Acc}} & \multirow{2}{*}{\textbf{ACIM}} & {\textbf{2-Step}} & \multicolumn{2}{c}{\textbf{L1-norm-based pruning}} \\
\cline{5-6}

&&& {\textbf{Pruning}} & {\textbf{On Edge}} & {\textbf{On Cloud}}\\ 
\hline
VGG16 & 77\% & {\textbf{20ms}} & 25ms & 51ms & 236ms\\
VGG19 & 83\% & {\textbf{34ms}} & 50ms & 86ms & 238ms\\
ResNet34 & 88\% & 67ms & \textbf{65ms} & 69ms & 237ms\\
\bottomrule 
\end{tabular}}
\label{tab01}
\vspace{-0.3cm}
\end{table}

{\color{black}As shown in Table \ref{tab01}, we compared our proposed ACIM method with 2-step pruning method and l1-norm-based pruning method. L1-norm-based pruning method is a traditional method and both ACIM method and 2-step pruning method are designed for collaborative inference. We applied these methods to prune three types of recognition model, and ensured that each model, after pruning by different methods and 5 epochs of fine-tuning, attained equivalent accuracy. Then, we evaluated the collaborative inference latency of the model pruned by our ACIM method and 2-step pruning method, and tested the inference latency on the edge of the model pruned by l1-norm based pruning method and on the cloud server with image data transmission. The results demonstrate that our proposed method achieved the lowest inference latency, showcasing a reduction of up to 32\% compared with 2-step pruning method and a reduction of up to 60\% compared with l1-norm based pruning method.}

\section{Conclusion}
In this paper, we propose the E-SC3I framework, establishing four mechanisms: emergency computing, emergency caching, integrated communication and sensing, and intelligence empowerment. These mechanisms ensure the transmission of command information while adaptively adjusting the network deployment plan based on the sensing of the time-varying emergency environment. This ensures a rapid, reliable, and sustained emergency communication dedicated network. Furthermore, we propose the ACIM method for emergency computing, utilizing a hierarchical reinforcement learning approach to optimize both the pruning and partition strategy of a recognition model. Experimental results demonstrate that this method can effectively reduce collaborative inference latency while ensuring inference accuracy. In the future, we plan to continue improving the proposed E-SC3I framework and provide more technical details.


\bibliographystyle{IEEEtran}
\bibliography{reference}

\begin{thebibliography}{10}
\providecommand{\url}[1]{#1}
\csname url@samestyle\endcsname
\providecommand{\newblock}{\relax}
\providecommand{\bibinfo}[2]{#2}
\providecommand{\BIBentrySTDinterwordspacing}{\spaceskip=0pt\relax}
\providecommand{\BIBentryALTinterwordstretchfactor}{4}
\providecommand{\BIBentryALTinterwordspacing}{\spaceskip=\fontdimen2\font plus
\BIBentryALTinterwordstretchfactor\fontdimen3\font minus \fontdimen4\font\relax}
\providecommand{\BIBforeignlanguage}[2]{{%
\expandafter\ifx\csname l@#1\endcsname\relax
\typeout{** WARNING: IEEEtran.bst: No hyphenation pattern has been}%
\typeout{** loaded for the language `#1'. Using the pattern for}%
\typeout{** the default language instead.}%
\else
\language=\csname l@#1\endcsname
\fi
#2}}
\providecommand{\BIBdecl}{\relax}
\BIBdecl

\bibitem{b2}
L.~Wang, Y.~Ai, N.~Liu, and A.~Fei, ``User association and resource allocation in full-duplex relay aided noma systems,'' \emph{IEEE Internet of Things Journal}, vol.~6, no.~6, pp. 10\,580--10\,596, 2019.

\bibitem{b3}
L.~Xu, Z.~Yang, H.~Wu, Y.~Zhang, Y.~Wang, L.~Wang, and Z.~Han, ``Socially driven joint optimization of communication, caching, and computing resources in vehicular networks,'' \emph{IEEE Transactions on Wireless Communications}, vol.~21, no.~1, pp. 461--476, 2022.

\bibitem{b13}
W.~Shi, Y.~Hou, S.~Zhou, Z.~Niu, Y.~Zhang, and L.~Geng, ``Improving device-edge cooperative inference of deep learning via 2-step pruning,'' in \emph{IEEE INFOCOM 2019 - IEEE Conference on Computer Communications Workshops (INFOCOM WKSHPS)}, 2019, pp. 1--6.

\bibitem{b17}
L.~Wang, J.~Zhang, J.~Chuan, R.~Ma, and A.~Fei, ``Edge intelligence for mission cognitive wireless emergency networks,'' \emph{IEEE Wireless Communications}, vol.~27, no.~4, pp. 103--109, 2020.

\bibitem{b5}
Q.~Wei, L.~Wang, L.~Xu, Z.~Tian, and Z.~Han, ``Hierarchical coded caching for multiscale content sharing in heterogeneous vehicular networks,'' \emph{IEEE Transactions on Vehicular Technology}, vol.~71, no.~6, pp. 5770--5786, 2022.

\bibitem{b6}
Y.~Shi, K.~Yang, T.~Jiang, J.~Zhang, and K.~B. Letaief, ``Communication-efficient edge ai: Algorithms and systems,'' \emph{IEEE Communications Surveys \& Tutorials}, vol.~22, no.~4, pp. 2167--2191, 2020.

\bibitem{b8}
Y.~He, J.~Lin, Z.~Liu, H.~Wang, L.-J. Li, and S.~Han, ``Amc: Automl for model compression and acceleration on mobile devices,'' in \emph{Proceedings of the European conference on computer vision (ECCV)}, 2018, pp. 784--800.

\bibitem{b11}
J.~Yim, D.~Joo, J.~Bae, and J.~Kim, ``A gift from knowledge distillation: Fast optimization, network minimization and transfer learning,'' in \emph{Proceedings of the IEEE conference on computer vision and pattern recognition}, 2017, pp. 4133--4141.

\bibitem{b12}
B.~Jacob, S.~Kligys, B.~Chen, M.~Zhu, M.~Tang, A.~Howard, H.~Adam, and D.~Kalenichenko, ``Quantization and training of neural networks for efficient integer-arithmetic-only inference,'' in \emph{Proceedings of the IEEE conference on computer vision and pattern recognition}, 2018, pp. 2704--2713.

\bibitem{b14}
Y.~Kang, J.~Hauswald, C.~Gao, A.~Rovinski, T.~Mudge, J.~Mars, and L.~Tang, ``Neurosurgeon: Collaborative intelligence between the cloud and mobile edge,'' \emph{ACM SIGARCH Computer Architecture News}, vol.~45, no.~1, pp. 615--629, 2017.

\bibitem{b15}
J.~H. Ko, T.~Na, M.~F. Amir, and S.~Mukhopadhyay, ``Edge-host partitioning of deep neural networks with feature space encoding for resource-constrained internet-of-things platforms,'' in \emph{2018 15th IEEE International Conference on Advanced Video and Signal Based Surveillance (AVSS)}, 2018, pp. 1--6.

\bibitem{b16}
J.~Shao and J.~Zhang, ``Bottlenet++: An end-to-end approach for feature compression in device-edge co-inference systems,'' in \emph{2020 IEEE International Conference on Communications Workshops (ICC Workshops)}, 2020, pp. 1--6.

\end{thebibliography}

\end{document}